\documentclass[journal]{IEEEtran}
\ifCLASSINFOpdf
\else
\fi
\usepackage{times}
\usepackage{epsfig}
\usepackage{graphicx}
\usepackage{amsmath}
\usepackage{amssymb}
\usepackage{subfigure}
\usepackage{multirow}
\usepackage{color}
\usepackage{ragged2e}
\usepackage[noend]{algpseudocode}
\usepackage{algorithmicx,algorithm}

\begin{document}
\title{Attribute-Guided Network for Cross-Modal Zero-Shot Hashing}

\author{Zhong Ji,
        Yuxin Sun,
        Yunlong Yu,
        Yanwei Pang,
        Jungong Han

\thanks{Z.~Ji, Y.~Sun, Y.~Yu, and  Y.~ Pang are with the School of Electrical and Information Engineering, Tianjin University, Tinjin 300072, China (e-mail: \{jizhong,~sunyuxin,~yuyunlong,~pyw\}@tju.edu.cn).}% <-this %
%\thanks{$^\ddag$Part of this work was done while Yunlong Yu was at State University of
%	New York, Binghamton University.}% stops a space
\thanks{J. Han is with the School of Computing \& Communications, Lancaster University, UK (e-mail: jungong.han@lancaster.ac.uk).}
}

%\markboth{Journal of \LaTeX\ Class Files,~Vol.XXX, No.XXXX, Nov.~XXXX}%

\maketitle

\begin{abstract}
Zero-Shot Hashing aims at learning a hashing model that is trained only by instances from seen categories but can generate well to those of unseen categories. Typically, it is achieved by utilizing a semantic embedding space to transfer knowledge from seen domain to unseen domain. Existing efforts mainly focus on single-modal retrieval task, especially Image-Based Image Retrieval (IBIR). However, as a highlighted research topic in the field of hashing, cross-modal retrieval is more common in real world applications. To address the Cross-Modal Zero-Shot Hashing (CMZSH) retrieval task, we propose a novel Attribute-Guided Network (AgNet), which can perform not only IBIR, but also Text-Based Image Retrieval (TBIR). In particular, AgNet aligns different modal data into a semantically rich attribute space, which bridges the gap caused by modality heterogeneity and zero-shot setting. We also design an effective strategy that exploits the attribute to guide the generation of hash codes for image and text within the same network. Extensive experimental results on three benchmark datasets (AwA, SUN, and ImageNet) demonstrate the superiority of AgNet on both cross-modal and single-modal zero-shot image retrieval tasks.
\end{abstract}

\IEEEpeerreviewmaketitle

\section{Introduction}

\IEEEPARstart{R}{cently}, hashing-based multimedia retrieval approaches have attracted a lot of attention, mainly owing to their fast retrieval speed and low storage cost \cite{shao16tc}, \cite{shao17ijcv}, \cite{han17tc}. Generally, these approaches fall into two categories: unsupervised hashing \cite{shao16tc}, \cite{shao17ijcv}, \cite{shen15tip}, \cite{gong13pami_itq} and supervised hashing \cite{shen15cvpr_sdh}, \cite{liu12cvpr}. The former usually applies the statistics information, such as manifold structure \cite{shen15tip} and the variance of feature \cite{gong13pami_itq}, to generate the hash function with the intention to preserve the similarity space, while the latter explores the semantic supervision information, \textit{e.g.}, class label, to capture the intrinsic property of data. Because more knowledge is utilized, supervised hashing approaches usually achieve better performance than those of unsupervised ones. However, one deficiency of supervised hashing approaches is that a large number of labeled instances are required for training the model, which is time-consuming and labor-intensive. In addition, it is very difficult to annotate sufficient training data for the new concepts in a timely manner, and also, impractical to retrain the hashing model whenever the retrieval system meets a new concept \cite{yang16mm}.

\begin{figure}[!ht]
	\centering
	\includegraphics[width=0.95\linewidth,height=0.50\linewidth]{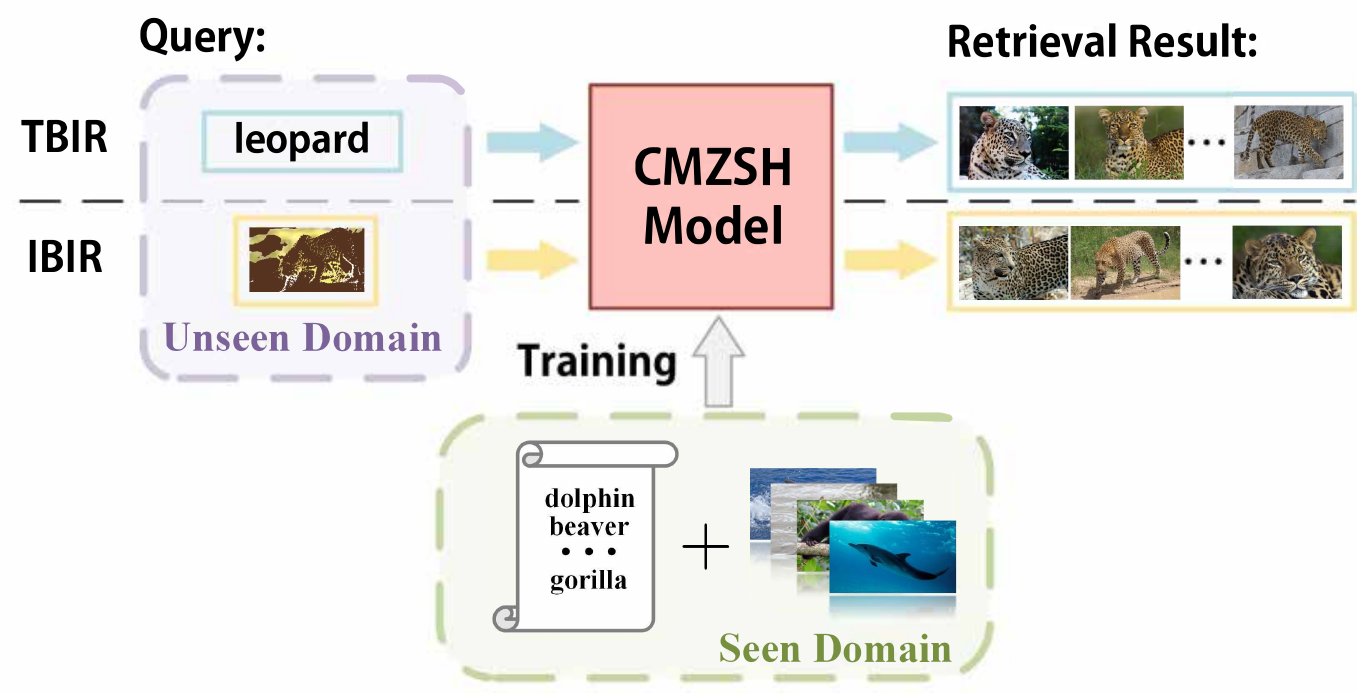}
	\caption{An illustration of Cross-Modal Zero-Shot Hashing (CMZSH). Typically, a CMZSH model is trained by texts and images in seen domain. At testing stage, the CMZSH model mainly tackles two tasks in unseen domain, \textit{i.e.,} Text-Based Image Retrieval (TBIR) and Image-Based Image Retrieval (IBIR). For TBIR, the query set are texts and the retrieval set are images. For IBIR, both the query and the retrieval sets are images.}
	\setlength{\belowcaptionskip}{-0.2cm}
	
\end{figure}

To address this awkward situation, inspired by the success of Zero-Shot Learning (ZSL) \cite{Lampert14pami}, \cite{fu14pami}, \cite{yang17cvpr}, \cite{ji17is}, Zero-Shot Hashing (ZSH) is developed recently \cite{yang16mm}, \cite{han17ijcai_sitnet}. Its goal is to encode images of unseen categories with the hash funciton trained by only those of seen categories by incorporating the ideas of supervised hashing approaches and ZSL. Transferring Supervised Knowledge (TSK) \cite{yang16mm} is the pioneering method in ZSH. The authors propose to employ the semantic vectors as a bridge to transfer available supervision information from seen categories to unseen categories. Further, Guo \textit{et al}. \cite{han17ijcai_sitnet} present a deep ZSH method, named Similarity Transfer Network (SitNet). Specifically, SitNet applies a multi-task architecture to leverage the supervision knowledge of seen categories and the semantic vectors simultaneously, and employs a straight-through estimator to avoid information loss caused by real-value relaxation. Although these methods have achieved impressive performance, there is still a serious limitation for them. That is, the existing ZSH approaches only focus on Image-Based Image Retrieval (IBIR) task, where both the query and the retrieval sets are images. In fact, Text-Based Image Retrieval (TBIR), \textit{i.e.}, leveraging textual description to search images, is also very common in the real-life scenario.

The aforementioned limitation motivates us to consider investigating ZSH in a cross-modal retrieval setting, which we call Cross-Modal Zero-Shot Hashing (CMZSH). Specifically, CMZSH mainly deals with two different tasks, one is IBIR, and the other is TBIR. That is to say, CMZSH broadens the scope of conventional ZSH from single-modal application to cross modal application. An illustration is described in Fig. 1. It should be noted that Image-Based Text Retrieval (IBTR) also belongs to the scope of Cross-Modal Hashing. However, since only one category name is corresponding to a class of images for most popular ZSH datasets, IBTR in this situation actually degenerates into a ZSL (also called zero-shot image classification) problem, which is not the focus of our work.

To achieve CMZSH, the following challenges should be addressed. 1) Modality heterogeneity. As query set and retrieval set are likely to be from different modalities, the generated hash codes are expected to have an additional property that preserves the semantic relationship between both modalities. 2) Category migration. It is an inherent problem of ZSL that the learning model should have the ability of handling the instances from unseen categories. Therefore, CMZSH needs to exploit the transferable knowledge that bridges the gap between seen categories and unseen categories. 3) Semantic similarity preservation. The hash function is actually a projection from high dimensional real-value features to low dimensional binary space. To implement effective nearest neighbor search, the generated binary hash codes are necessary to inherit the semantic similarity relationship of high dimensional real-value features.

In this paper, we address the above issues with the proposed Attribute-Guided Network (AgNet) framework. Specifically, to narrow the semantic gap brought by modality heterogeneity and category migration, we map both the visual features and the textual features into a common space, respectively. In this work, we utilize the class-level attribute space as the common space. In this way, the two different modalities are aligned into a high-level semantic space. Using the embeddings of different modalities in the attribute space as inputs, both visual and textual hash codes are obtained from a shared deep neural network. Besides, the relationships between different modalities are constructed via a category similarity matrix formulated with the pair-wise class label. Moreover, to preserve the relationship of different categories, attribute similarity is further introduced to restrict the distances of different categories in the same modality. 

We summarize our highlights as below:
\begin{enumerate}
	\item We address the cross-modal retrieval problem in ZSH, \textit{i.e.}, Cross-Modal Zero-Shot Hashing (CMZSH), via a novel deep hashing neural network. It can perform not only IBIR, but also TBIR. To the best of our knowledge, it is the first work to study the cross-modal hashing retrieval in the zero-shot setting.
	\item By exploiting the class-level attributes information, we propose an Attribute-Guided Network (AgNet) framework. It first maps two different modalities into a common attribute space, which acts as a hub to bridge unseen and seen categories, as well as visual and textual modalities. Then, an effective strategy is designed to generate two individual hash codes for image and text within the same network. Specifically, we exploit the attribute to guide the generation of hash codes by preserving the category similarity and attribute similarity.
	\item The experimental results for both IBIR and TBIR tasks on three popular benchmark datasets demonstrate that the proposed AgNet achieves competitive performance.
\end{enumerate}

\section{Related Work}
In this section, we will introduce some research progresses closely related to our work, including cross-modal hashing and zero-shot hashing. In fact, CMZSH can be viewed as a special case for them. CMZSH also falls into the domains of hashing-based retrieval and zero-shot learning. Due to the limited space, please refer to \cite{liu16ieee} and \cite{fu17magazine} for more elaborate surveys about them.
\subsection{Cross-Modal Hashing}
Cross-Modal Hashing (CMH) is a widely used retrieval technique \cite{han17tc}, \cite{ding16tip}, \cite{zhang14aaai}, most of which tackle the problems of Text-Based Image Retrieval (TBIR) and Image-Based Text Retrieval (IBTR). This is usually implemented by generating two respective hash codes for each individual modality. In this way, different modalities can be computed in the same hashing space. A number of methods have been proposed, which can be generally divided into two categories: unsupervised methods and supervised methods. As one of the representative unsupervised cross-modal methods, Collective Matrix Factorization Hashing (CMFH) \cite{ding16tip} generates cross-modal hash codes in a latent semantic space shared by both modalities via collective matrix factorization technique. To explore the heterogeneous correlation in different modalities, Liu \textit{et al.} \cite{liu17cvpr} propose a novel CMH scheme using fusion similarity from the multiple modalities. 

On the other hand, supervised CMH methods usually perform better than unsupervised ones since they can fully utilize the intrinsic property in data. For example, Zhang \textit{et al.} \cite{zhang14aaai} merge semantic labels into hashing learning procedure and propose a Semantic Correlation Maximization Hashing (SCMH) method. Lin \textit{et al.} \cite{lin15cvpr} convert the semantic similarity of instances into a probability distribution and generate hash codes by minimizing the KL-divergence. Liu \textit{et al.} \cite{liu17arxiv} propose a graph-regularized Supervised Matrix Factorization Hashing (SMFH) framework with a collective non-negative matrix factorization technique. With the renaissance of the deep neural network, deep learning has proved its outperformance in this field. For instance, Jiang \textit{et al.} \cite{jiang17cvpr} first propose an end-to-end deep neural network framework to address the CMH problem. However, they just utilize the inter-modal relationship but ignore intra-modal information. To address this problem, Yang \textit{et al.}\cite{yang17aaai} use pairwise labels to exploit intra-modal similarity and propose a Pairwise Relationship Guided Deep Hashing (PRDH) method.

Our proposed CMZSH framework follows the idea of supervised CMH, which leverages semantic supervision information to generate different hashing codes for each modality to ensure they are able to interact with each other. However, different from CMH, CMZSH has to tackle an additional zero-shot problem. That is, the supervision knowledge is limited to seen categories, which is the only information in learning reliable hash function for transforming modalities of unseen categories into binary codes. Therefore, CMZSH is more challenging.

\subsection{Zero-Shot Hashing}
Zero-Shot Hashing (ZSH) is a marriage of zero-shot learning and hashing-based retrieval techniques. It is proposed to tackle the close-set limitation in hashing-based retrieval approaches, \textit{i.e}., the concepts of possible testing instances in either dataset or query set are within the training set \cite{yang16mm}, \cite{han17ijcai_sitnet}. Therefore, ZSH explores only the information from seen categories to build hash functions to retrieve the images in unseen categories. 
\begin{figure*}[!t]
	\centering
	\includegraphics[width=0.96\linewidth, height=0.29\linewidth]{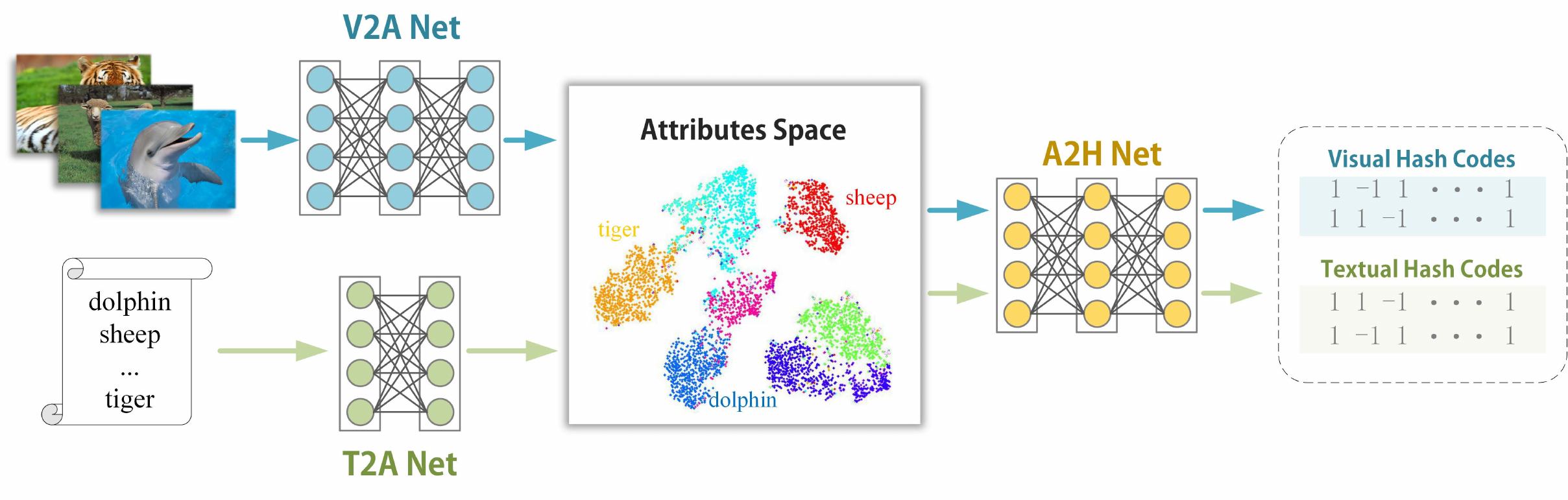}
	\setlength{\abovecaptionskip}{-0.05cm}
	\caption{Architecture overview of the proposed AgNet approach. It consists of two stages. First, V2A Net and T2A Net embed the inputs of image and text into a shared attribute space, respectively. Next, A2H Net encodes the visual and textual vectors in attribute space into visual hash codes and textual hash codes, respectively. The shared attribute space enables the knowledge transferability from seen categories to unseen categories. And the A2H Net makes the cross-modal retrieval feasible.}
	\setlength{\belowcaptionskip}{-0.4cm}
\end{figure*}

As an emerging research topic, the existing ZSH methods mainly focus on IBIR task. For example, in the pioneering work proposed by Yang \textit{et al.} \cite{yang16mm}, the labels of each seen category are converted into semantic embedding representations via word2vec model \cite{wordvec}, by which the supervision knowledge in seen categories can be transferred to unseen ones. Then, hash codes are generated by projecting the visual representation to the embedding space. Instead of using word vector as semantic representation in \cite{yang16mm}, Xu \textit{et al.} \cite{xu17icme} adopt semantically-rich attribute information as transferable knowledge. Further, Guo \textit{et al.} \cite{han17ijcai_sitnet} propose a multi-task framework to simultaneously exploit the supervision information from visual concepts and semantic representations. Specifically, they leverage the hash codes to capture the semantic similarity relationship in a transferable semantic embedding space and propose a center regularization loss to preserve both intra-concept similarity and inter-concept distance. In addition, under the transductive setting \cite{fu15pami}, \cite{yu17tnn}, Lai \textit{et al.} \cite{lai17arxiv} propose a transductive zero-shot hashing method via coarse-to-fine similarity mining. In this way, a greedy binary classification network is first used to detect the most informative images from unseen category images. After that, the fine similarity mining module further finds the similarities among the informative images. However, since these ZSH approaches are designed for IBIR task, they have a natural deficiency that cannot encode the text into hash codes. Therefore, they are hardly applied for TBIR.

To achieve CMZSH, the idea of ZSH should be combined with that of CMH. This is exactly what this paper is going to tackle.

\begin{table}[!ht]
	\centering
	\caption{\upshape The main notations.}
	\begin{tabular}{c|c}
		\hline
		Notation & Description \\
		\hline
		$N$ & number of instances\\
		$s$ & number of seen categories\\
		$u$ & number of unseen categories\\
		$d$ & number of attributes\\
		$c$ & hash codes length\\
		$l$ & dimensionality of visual space\\
		$k$ & dimensionality of textual space\\
		$\mathbf{x}\in \mathbb{R}{^l}$ & visual representation vector\\
		$\mathbf{z}\in \mathbb{R}{^k}$ & textual representation vector\\
		$\mathbf{y}\in \mathbb{R}{^{s+u}}$ & label vector\\
		$\mathbf{a}\in \mathbb{R}{^d}$ & ground-truth attribute vector\\
		${\bf{\hat a}}^{\left( v \right)}\in \mathbb{R}{^d}$ &  predicted attribute vector in visual modality \\
		${\bf{\hat a}}^{\left( t \right)}\in \mathbb{R}{^d}$ &  predicted attribute vector in textual modality \\
		${\bf{S}}^{\left( c \right)}\in \mathbb{R}{^{n \times n}}$ & category similarity matrix\\
		${\bf{S}}^{\left( att \right)}\in \mathbb{R}{^{n \times n}}$ & attribute similarity matrix\\		
		${\bf{P}}\in \mathbb{R}{^{c \times n}}$ & outputs matrix of A2H Net in visual modality \\
		${\bf{Q}}\in \mathbb{R}{^{c \times n}}$ & outputs matrix of A2H Net in textual modality \\
		${\bf{B}}\in \mathbb{R}{^{c \times n}}$ & hash codes matrix \\	
		\hline

	\end{tabular}
	%	\vspace{-0.4cm}
\end{table}
\section{The proposed AgNet algorithm}
\subsection{Problem Definition}
In order to address the CMZSH problem, both the requirements of knowledge transferability from seen categories to unseen categories and cross-modal retrieval should be fulfilled. Attributes and word vectors  are two most popular side information in ZSL \cite{lampert09cvpr}, \cite{akata13cvpr}, \cite{word16eccv}, \cite{rohrbach10cvpr}. Specifically, attributes define a few properties of an object, such as color, shape, and the presence or absence of a certain body part. They are shared across both seen and unseen categories. Word vectors represent words as vectors based on distributed language representation techniques, and theoretically, they can encode arbitrary concepts into sematic vectors. Therefore, both attributes and word vectors can construct a semantic space to transfer the knowledge from seen categories to unseen categories, meaning either of them can be selected as the candidate semantic space in CMZSH. Further, different approaches can be designed to generate both visual and textual hash codes from either attributes or word vectors space, which enables the cross-modal retrieval feasible. In this paper, we only focus on the usage of attributes. That is to say, we exploit attributes as the intermediary space, from which the hash codes are encoded.

Suppose we are given $N$ training instances ${{\cal D}_{tr}} = \{ {d_i} = ({{\bf{x}}_i},{{\bf{z}}_i} ,{{\bf{y}}_i})  {\kern 2pt},i= 1, ... ,N\} $ from $s$ labeled seen categories $S  =  \{ 1,2, ... s\} $, where ${{\bf{x}}_i} \in \mathbb{R}{^l}$ is the visual representation, ${{\bf{z}}_i} \in \mathbb{R}{^k}$ is the textual semantic representation of its corresponding category name and ${{\bf{y}}_i} \in {\{ 0,1\} ^s}$ is the label vector represented as one-hot encoding. Note that the different modalities mainly refer to image and text in this paper. Besides, each instance is also annotated with a binary attribute vector denoted as ${{\bf{a}}_i} \in {\{ 0, 1\} ^d}$. Under the zero-shot setting, there also exist unseen categories $U  = \{ s + 1, ...,s + u\} $, which is disjoint with the labeled seen categories, \textit{i.e.}, $S \cap U = \emptyset $.

\subsection{Network Architecture}
The overall framework of the proposed AgNet framework is illustrated in Fig. 2. It consists of three components: \expandafter{\romannumeral1}) \textbf{V2A Net}. The output of penultimate layer (before the Softmax layer) of the fine-tuned GoogleNet \cite{googlenet} is first extracted as the visual features. After that, these CNN features are utilized as the input to a deep neural network with three fully-connected layers, which embeds the visual features to attribute vectors. \expandafter{\romannumeral2}) \textbf{T2A Net}. We use word2vector model \cite{wordvec} to represent the text input, which has been trained on the Wikipedia corpus. It is a 1000-dimensional vector for each category name. T2A Net is a two-layer neural network that is used to establish the word vectors to attributes projection. \expandafter{\romannumeral3}) \textbf{A2H Net}. Unlike the existing deep cross-modal hashing methods that generate hash codes from two independent networks (one for image, and the other for text), AgNet accomplishes the hash codes generation only with a single three-layer neural network. Specifically, it utilizes the predicted attribute vectors (or called attribute embedding vectors) as input, and outputs both visual and textual hash codes. Table \uppercase\expandafter{\romannumeral2} shows the configuration of AgNet. It needs to be highlighted that the architecture of the neural network is not the focus of this work, what we want to prove is that attribute-guide framework is reasonable and beneficial for the performance of CMZSH.

\begin{table}[!t]
	\centering
	\caption{\upshape The network architecture details of the proposed AgNet. ``Full''denotes the fully-connected layer, ``Relu'' and ``Sigmoid'' denote activation functions.}
	\begin{tabular}{|c|c|c|}
		\hline
		Sub-network & Layer & Configuration \\
		\hline
		\hline
		\multirow{3}{*}{V2A Net} & Full1+Relu & 1024 \\
%		\cline{2-3}
		& Full2+Relu & 512 \\
%		\cline{2-3}
		& Full3+Sigmoid & number of attributes $d$\\
		\hline
		\multirow{2}{*}{T2A Net} & Full1+Relu & 1000 \\
%		\cline{2-3}
		& Full2+Sigmoid & number of attributes $d$\\
		\hline
		\multirow{3}{*}{A2H Net} & Full1+Relu & 128 \\
%		\cline{2-3}
		& Full2+Sigmoid & 128 \\
%		\cline{2-3}
		& Full3 & hash codes length $c$\\
		\hline
	\end{tabular}
	\vspace{-0.2cm}
\end{table}

\subsection{Objective Function}
We first design the objective functions for the V2A Net and the T2A Net, whose purpose is to transform the inputs of image and text to the attribute space. Their transformation functions are denoted as ${f_v}$ and ${f_t}$, respectively. Let ${\bf{\hat a}}_i^{\left( v \right)} = {f_v}\left( {{{\bf{x}}_i}} \right) \in \mathbb{R}{^d}$ denote the predicted attribute vector of each visual representation ${{\bf{x}}_i}$ while ${\bf{\hat a}}_i^{\left( t \right)} = {f_t}\left( {{{\bf{z}}_i}} \right) \in \mathbb{R}{^d}$ denotes the predicted attribute vector of each textual representation ${{\bf{z}}_i}$. Given a training set of instances and their corresponding category attribute vectors, the V2A Net and the T2A Net are both trained with the cross-entropy objective function:
\begin{equation}
\begin{aligned}
{{\rm{{\cal L}}}_{att}} =  - \frac{1}{N}\sum\limits_{i = 1}^N {{{\mathbf{a}}_i}^T\log \left( {{{{\mathbf{\hat a}}}_i}} \right)  + {{\left( {{\mathbf{1}} - {{\mathbf{a}}_i}} \right)}^T} \log (1 - {{{\bf{\hat a}}}_i})} ,
\end{aligned}
\end{equation}
where ${{\bf{a}}_i}$ denotes the attribute vector, and ${{\bf{\hat a}}_i}$ is the predicted attribute vector ${{\mathbf{\hat a}^{\left(v\right)}}_i}$ for V2A Net or ${{\mathbf{\hat a}^{\left(t\right)}}_i}$ for T2A Net. This objective function ensures that the predicted attribute vectors approximate to the distribution of original attribute vectors.

Then, the key challenge is how to realize the purpose of A2H Net, \textit{i.e.,} to generate two individual hash codes for image and text from the attribute space. We design three losses to achieve this purpose: i) category similarity loss; ii) attribute similarity loss; and iii) regularization loss.

The category similarity loss is proposed to ensure the different predicted attribute vectors of different modalities in the same category can generate similar hash codes, while those in different categories have distinct differences. Given the predicted attribute vectors ${\bf{\hat a}}_{}^{\left( v \right)}$ and ${\bf{\hat a}}_{}^{\left( t \right)}$ of image and text, respectively,  denote ${\mathbf{P}_{*i}} = g\left( {{\bf{\hat a}}_i^{^{\left( v \right)}};\theta } \right) \in \mathbb{R}{^c}$ and ${\mathbf{Q}_{*i}} = g\left( {{\bf{\hat a}}_i^{^{\left( t \right)}};\theta } \right) \in \mathbb{R}{^c}$ as their outputs of A2H Net, where $g$ is the transformation function for A2H Net and $\theta$ is the parameters for it. Moreover, use ${\mathbf{\Theta} _{ij}} = \frac{1}{2}{\mathbf{P}_{*i}}^T{\mathbf{Q}_{*j}}$ to represent  the neighbor relationship between $\mathbf{P}_{*i}$ and $\mathbf{Q}_{*j}$ in the Hamming space . Denote $\mathbf{S}_{}^{(c)} \in \mathbb{R}{^{n \times n}}$as the category similarity, where $\mathbf{S}_{ij}^{(c)} = 1$ when ${\mathbf{y}_i} = {\mathbf{y}_j}$ and $\mathbf{S}_{ij}^{(c)} = 0$ otherwise. By using the negative log likelihood of the inter-modal similarities, we formulate the category similarity loss as:
\begin{equation}
\begin{aligned}
{{\rm{{\cal L}}}_{cs}} =  - \sum\limits_{i,j = 1}^N {\left( {\mathbf{S}_{ij}^{(c)}{\mathbf{\Theta} _{ij}} - \log \left( {1 + {e^{{\mathbf{\Theta} _{ij}}}}} \right)} \right)}.
\end{aligned}
\end{equation}

By minimizing this objective function, the Hamming distances for those instances within the same category but with different modalities are reduced, whereas the distances are getting larger for those with different categories. Therefore, the category similarity is preserved between different modalities.

In addition, an effective hash code should also be equipped with the discriminative ability in a single modality. Hence, attribute similarity matrix $\mathbf{S}_{}^{\left( {att} \right)}$ is introduced to make the intra-modal hash codes more discriminable. Let $\textbf{S}_{ij}^{\left( {att} \right)} = \cos \left( {{{\bf{a}}_i},{{\bf{a}}_j}} \right) - \textbf{S}_{ij}^{(c)}$, where $\cos \left( {{{\bf{a}}_i},{{\bf{a}}_j}} \right)$ is the cosine distance between attribute vectors of ${{\bf{a}}_i}$ and ${{\bf{a}}_j}$. $\textbf{S}_{}^{\left( {att} \right)}$is a modified cosine similarity, which is used to measure the semantic similarities among different categories. Different from the binary label similarity $\mathbf{S}_{}^{\left( c \right)}$,  $\mathbf{S}_{}^{\left( {att} \right)}$ utilizes a real-value to describe the similarities among different categories. It is used in the attribute similarity loss ${{\rm{{\cal L}}}_{as}}$ as a guide for the generation of visual hash codes. Specifically, if two attribute vectors from different categories are similar, their corresponding visual instances should be given a higher penalty such that their hash codes have higher discriminative ability. If two instances are from the same category, we do not give them penalty, that is why $\mathbf{S}_{}^{\left( c \right)}$ is subtracted. The attribute similarity loss is defined as follow:
\begin{equation}
{{\rm{{\cal L}}}_{as}} = \sum\limits_{i,j = 1}^N {\sigma \left( {{\phi _{i,j}} \mathbf{S}_{i,j}^{\left( {att} \right)}} \right)},
\end{equation}
where ${\mathbf{\phi} _{ij}} = \frac{1}{2}\mathbf{P}_{*i}^T{\mathbf{P}_{*j}}$ represents the neighbor relationship of $\mathbf{P}_{*i}$ and $\mathbf{P}_{*j}$ in Hamming space, and $\sigma \left( \bullet \right)$ denotes the sigmoid function. The sigmoid function is applied to restrict the scope of this term. By minimizing this term, those instances closed in the attribute space and from different categories will be uncoupled in the Hamming space.

Meanwhile, we use $\left\| {\mathbf{B} - \mathbf{P}} \right\|_F^2$ to make $\mathbf{P}$ approximate to hash codes. And $\left\| {{\mathbf{P}^T}{\bf{1}}} \right\|_F^2$ ensures each bit of the hash codes is balanced. Then, the regularization loss is formulated as follow:
\begin{equation}
{{\rm{{\cal L}}}_{reg}} = \left\| {\mathbf{B} - \mathbf{P}} \right\|_F^2 + \left\| {{\mathbf{P}^T}{\mathbf{1}}} \right\|_F^2,
\end{equation}
where $\mathbf{B} = sign\left( \mathbf{P} \right)$ , ${\bf{1}}$ denotes a vector with all elements being $1$.

Therefore, the overall objective function of the A2H Net is written as follow:
\begin{equation}
\begin{aligned}
{{\rm{{\cal L}}}_{A2H}}  &= {{\rm{{\cal L}}}_{cs}} + \lambda{{\rm{{\cal L}}}_{as}} + \eta{{\rm{{\cal L}}}_{reg}}\\
&=  - \sum\limits_{i,j = 1}^N {\left( {\mathbf{S}_{ij}^{(c)}{\mathbf{\Theta} _{ij}} - \log \left( {1 + {e^{{\mathbf{\Theta} _{ij}}}}} \right)} \right)}  \\ 
&+ \sum\limits_{i,j = 1}^N {\lambda\sigma \left( {{\mathbf{\phi} _{i,j}} \mathbf{S}_{i,j}^{\left( {att} \right)}} \right)}  + \eta\left(\left\| {\mathbf{B} - \mathbf{P}} \right\|_F^2 + \left\| {{\mathbf{P}^T}{\bf{1}}} \right\|_F^2\right),
\end{aligned}
\end{equation}
where $\lambda$ and $\eta$ are trade-off parameters to control the weight of each item.
\subsection{Optimization}
Our AgNet is trained in two steps. Firstly, V2A Net and T2A Net are separately learned with cross entropy functions. Then, using the predicted attribute vectors from two modalities, we train A2H Net according to Eq. (5). Back Propagation algorithm is adopted to optimize AgNet. For Eq.(5), the gradient of $\frac{{{{\rm{{\cal L}}}_{A2H}}}}{{\partial \mathbf{P}_{*i}^{}}}$ is calculated with:

\begin{equation}
\begin{aligned}
\frac{{{{\rm{{\cal L}}}_{A2H}}}}{{\partial \mathbf{P}_{*i}^{}}} 
&= \frac{1}{2}\sum\limits_{j = 1}^N {\left( {\sigma ({\mathbf{\Theta} _{ij}}) - \mathbf{S}_{ij}^{(c)}{\mathbf{Q}_{*j}}} \right)}  
\\&+ \frac{1}{2}\sum\limits_{j = 1}^N\lambda {\mathbf{P}_{*j}^{}\mathbf{S}_{i,j}^{\left( {att} \right)}\sigma ({\phi _{i,j}} \mathbf{S}_{i,j}^{\left( {att} \right)})(1 - \sigma ({\phi _{i,j}} \mathbf{S}_{i,j}^{\left( {att} \right)}))} \\
&+ 2\eta\left( {{\mathbf{P}_{*i}} - {\mathbf{B}_{*i}}} \right) + 2\eta\mathbf{P}_{*i}^T{\bf{1}}.
\end{aligned}
\end{equation}

Then, the gradient of weight in A2H Net can be calculated with $\frac{{{{\rm{{\cal L}}}_{A2H}}}}{{\partial P_{*i}^{}}}$ according to chain rule. The details of training A2H Net are shown in Algorithm 1. Using mini-batch Stochastic Gradient Descent algorithm, we fix the batch size to be 32. The initial learning rate is set as ${10^{ - 3}}$ and decreased by 0.01\% for each iteration. We choose the hyperparameter $\lambda$ and $\eta$ in AgNet according to the results on validation set and find the best performances can be achieved with $\lambda=\eta=1$. Therefore, we set $\lambda=\eta=1$. Our neural network is implemented with TensorFlow library on an NVIDIA 1080ti GPU server.
\begin{algorithm}[ht]
	\caption{Algorithm for training A2H Net}
	\hspace* {0.02in}{\bf Input:} 
	\\The predicted visual attribute vectors ${\mathbf{\hat{a}}}_{}^{\left( v \right)}$, \\
 the predicted textual attribute vectors ${\bf{\hat{a}}}_{}^{\left( t \right)}$, \\
	label matrix $\mathbf{Y}$ and attribute matrix $\mathbf{A}$ .\\
	\hspace*{0.02in} {\bf Output:} 
	\\Parameters $\theta $ in the A2H Net and binary codes $\mathbf{B}$.
	\begin{algorithmic}[1]
		\State {\textbf{Initialization:}
			Randomly initialize parameters $\theta $ of A2H Net, set mini-batch $M = 32$ and iteration number $l = \left\lfloor {{N \mathord{\left/
						{\vphantom {N M}} \right.
						\kern-\nulldelimiterspace} M}} \right\rfloor $}. % \State ºóÐ´Ò»°ãÓï¾ä

		\State{\textbf{Repeat}}
		\For{$iter = 1,2,...,l$} % For Óï¾ä£¬ÐèÒªºÍEndFor¶ÔÓ¦
		\State {Randomly sample $M$ instances.}
		\State {Calculate category similarity $\mathbf{S}_{}^{(c)}$.
			\State Calculate attribute similarity $\mathbf{S}_{}^{\left( {att} \right)}$.}
		\State Calculate $\mathbf{Q}$ and $\mathbf{P}$ by forward propagation.
		\State Get the corresponding binary code $\mathbf{B}$.
		\State Update the parameter $\theta $ by back propagation.
		\EndFor
		\State \textbf{until} a fixed number of iterations.
	\end{algorithmic}
\end{algorithm}

\begin{table*}[!ht]
	\centering
	\caption{\upshape Results on three benchmark datasets in Mean Average Precision (\%). The best results are marked in bold.}
	\begin{tabular}{|c|c|c|c|c|c|c|c|c|c|c|c|c|c|c|c|}
		\hline
		\multirow{2}{*}{Method} & \multicolumn{5}{c|}{AwA} & \multicolumn{5}{c|}{SUN} & \multicolumn{5}{c|}{ImageNet}\\
		\cline{2-16}
		& 8bits & 16bits & 32bits & 48bits & 64bits & 8bits & 16bits & 32bits & 48bits & 64bits & 8bits & 16bits & 32bits & 48bits & 64bits \\
		\hline
		\hline
		SCMH \cite{zhang14aaai} & 15.2 & 14.2 & 14.1 & 12.6 & 12.1 & 15.1 & 16.2 & 19.1 & 21.4 & 18.8 & 1.46 & 1.88 & 2.06 & 1.84 & 1.73 \\
		%		\hline
		SMFH \cite{liu17arxiv} & 17.7 & 19.3 & 21.5 & 22.9 & 21.6 & 12.6 & 12.1 & 12.4 & 12.6 & 13.1 & 1.38 & 1.33 & 2.00 & 2.23 & 2.40 \\
		%		\hline
		DCMH \cite{jiang17cvpr} & 11.9 & 9.8 & 12.7 & 9.8 & 10.3 & 12.3 & 12.6 & 13.7 & 13.5 & 14.1 & 1.00 & 1.04 & 1.03 & 1.00 & 1.01 \\
		%		\hline
		FSH \cite{liu17cvpr} & 12.7 & 14.1 & 14.2 & 12.6 & 12.1 & 19.7 & 20.8 & 16.2 & 18.7 & 16.5 & 1.44 & 1.95 & 2.31 & 2.65 & 2.72 \\
		%		\hline
		AgNet & \textbf{41.9} & \textbf{50.1} & \textbf{56.1} & \textbf{58.1} & \textbf{58.8}
	    & \textbf{21.1} & \textbf{21.3} & \textbf{23.5} & \textbf{24.5} & \textbf{26.6}
		& \textbf{3.80} & \textbf{5.26} & \textbf{5.89} & \textbf{5.98} & \textbf{5.77} \\
		\hline
	\end{tabular}
	%	\vspace{-0.4cm}
\end{table*}
\section{Experiment}
In this section, we implement both the single-model and cross-modal zero-shot retrieval tasks, i.e., IBIR and TBIR, on three benchmark datasets. And we compare the proposed AgNet approach with several existing state-of-the-art methods to demonstrate its effectiveness. 
\subsection{Datasets}
\textbf{Animals with Attributes (AwA)} \cite{lampert09cvpr}.  AwA dataset consists of 30,475 images from 50 animal categories and 85 associated class-level attributes. It is a popular dataset for ZSL. We follow the standard seen/unseen split \cite{lampert09cvpr}, where 40 categories with 24,295 images are taken as the seen domain and the remaining 10 categories with 6,180 images are adopted as the unseen domain.

\textbf{SUN attribute} \cite{patterson12cvpr_sun}. It is another widely used dataset in ZSL, which consists of 717 scene categories annotated by 102 attributes. Each category has 20 images, and there are totally 14,340 images. Following \cite{jayaraman14nips_sun_split}, we utilize 707 categories as the seen domain and the other 10 categories as the unseen domain.

\textbf{ImageNet} \cite{imagenet}. ImageNet is a large-scale image dataset organized according to the Word-Net \cite{wordnet}  hierarchy. As no attribute is annotated for this dataset, in our experiment, we use AwA dataset as an auxiliary dataset to construct the training set. Specifically, after removing 10 similar categories shared by two datasets\footnote{We eliminate 10 categories (\textit{i.e.}, dalmatian, collie, german shepherd, chihuahua, persian cat, siamese cat, bobcat, horse, deer, sheep) from AwA to construct the seen domain.}, we choose 40 categories with 21,832 images from AwA as seen domain and 100 animal categories with 129,622 images from ILSVRC2012 as the unseen domain.
\subsection{Cross-Modal Zero-Shot Hashing}
%\subsubsection{Benchmark Comparison}
Under cross-modal zero-shot retrieval setting, \textit{i.e.}, TBIR, the seen data are used for training the model. At the testing stage, the names of unseen categories are used as queries for retrieving images from the unseen domain. 

Since the existing ZSH approaches cannot tackle the cross-modal retrieval task, we choose the CMH approaches for comparison. Four existing state-of-the-art CMH approaches are selected for comparison, where SCMH \cite{zhang14aaai}, SMFH \cite{liu17arxiv}, and DCMH \cite{jiang17cvpr} are three representative supervised CMH methods, while FSH \cite{liu17cvpr} is an unsupervised CMH method. For all comparative approaches, we use the codes provided by the authors. As DCMH is an end-to-end CMH method, we utilize raw images as input. The others adopt the same visual features as ours, that is, the GoogleNet features \cite{googlenet} fine-tuned in the training set. Besides, we use the word2vec features \cite{wordvec} as textual features for all methods.

We use the Mean Average Precision (mAP) to evaluate the performances of the proposed AgNet and the comparative approaches. To observe the performance under different code length, we set the code length with 8, 16, 32, 48 and 64 bits , respectively. From the results shown in Table \uppercase\expandafter{\romannumeral3}, we have the following observations: \expandafter{\romannumeral1}) The proposed AgNet achieves the best performance on all three datasets consistently. All the comparative approaches have a relatively poor performance. This is mainly due to the reason that they are not designed for zero-shot settings, which leads to a worse generalization ability on the unseen domains. Specifically, it has a significant improvement on AwA dataset. For instance, the mAP performance of AgNet is 58.1\% with 48 bits, which has a 35.2\% absolute gain than that of the second best method SMFH. \expandafter{\romannumeral2}) The performances of AgNet on SUN dataset are inferior to those on AwA dataset. This is partly due to the fact that SUN is a fine-grained dataset in which there are few diversities in each category, making the learned hash codes be less discriminative. \expandafter{\romannumeral3}) AgNet also has a relatively small promotion on the large-scale ImageNet. Considering that the numbers of both the testing categories and instances in ImageNet are far more than those in AwA and SUN datasets, the improvements are still impressive. \expandafter{\romannumeral4}) The mAP performances of AgNet are positively related to code length in most situations, which indicates that the discriminative ability of hash codes increases with the growth of code length. By contrast, the mAP performances of comparative methods are unstable and without such a property.  In a word, the experimental results clearly demonstrate the superiority of the proposed AgNet approach in CMZSH task.

\subsection{Single-Modal Zero-Shot Hashing}
The existing ZSH methods mainly focus on single-modal retrieval, \textit{i.e.}, the query set and retrieval set are both constructed with the images. To evaluate the generalization of AgNet, we also implement AgNet in the single-modal ZSH task. As the scale of the unseen domain in SUN is insufficient to evaluate the performance of image retrieval task,  we just implement single-model retrieval on AwA and ImageNet datasets.

Following \cite{yang16mm} and \cite{han17ijcai_sitnet}, we randomly choose 10,000 instances from seen domain to construct the training set. As for testing, we randomly select 1,000 images from the unseen domain as the query set. The remaining unseen images and all seen domain images form the retrieval set.

We select the following state-of-the-art hashing methods as the baselines. IMH \cite{shen13cvpr_imh} and ITQ \cite{gong13pami_itq} are two representative unsupervised hashing methods, SDH \cite{shen15cvpr_sdh}, TSK \cite{yang16mm} and SitNet \cite{han17ijcai_sitnet} are three supervised hashing methods. In addition, TSK and SitNet are specially designed for zero-shot retrieval. The mAP and Precision within Hamming radius 2 are adopted as the evaluation metrics in this task. For all comparative approaches, we utilize GoogleNet features fine-tuned in the training set as the visual features. Following \cite{han17ijcai_sitnet}, we set the code length to be 8, 16, 32, and 48 bits, respectively. 
\begin{figure}[!h]
	\centering
	%	\setlength{\abovecaptionskip}{-0.05cm}
	%	\label{fig:fig3_a}
%	\includegraphics[width=0.80\linewidth,height=0.98\linewidth]{fig3_1.pdf}
		\includegraphics[width=0.8\linewidth,height=1.22\linewidth]{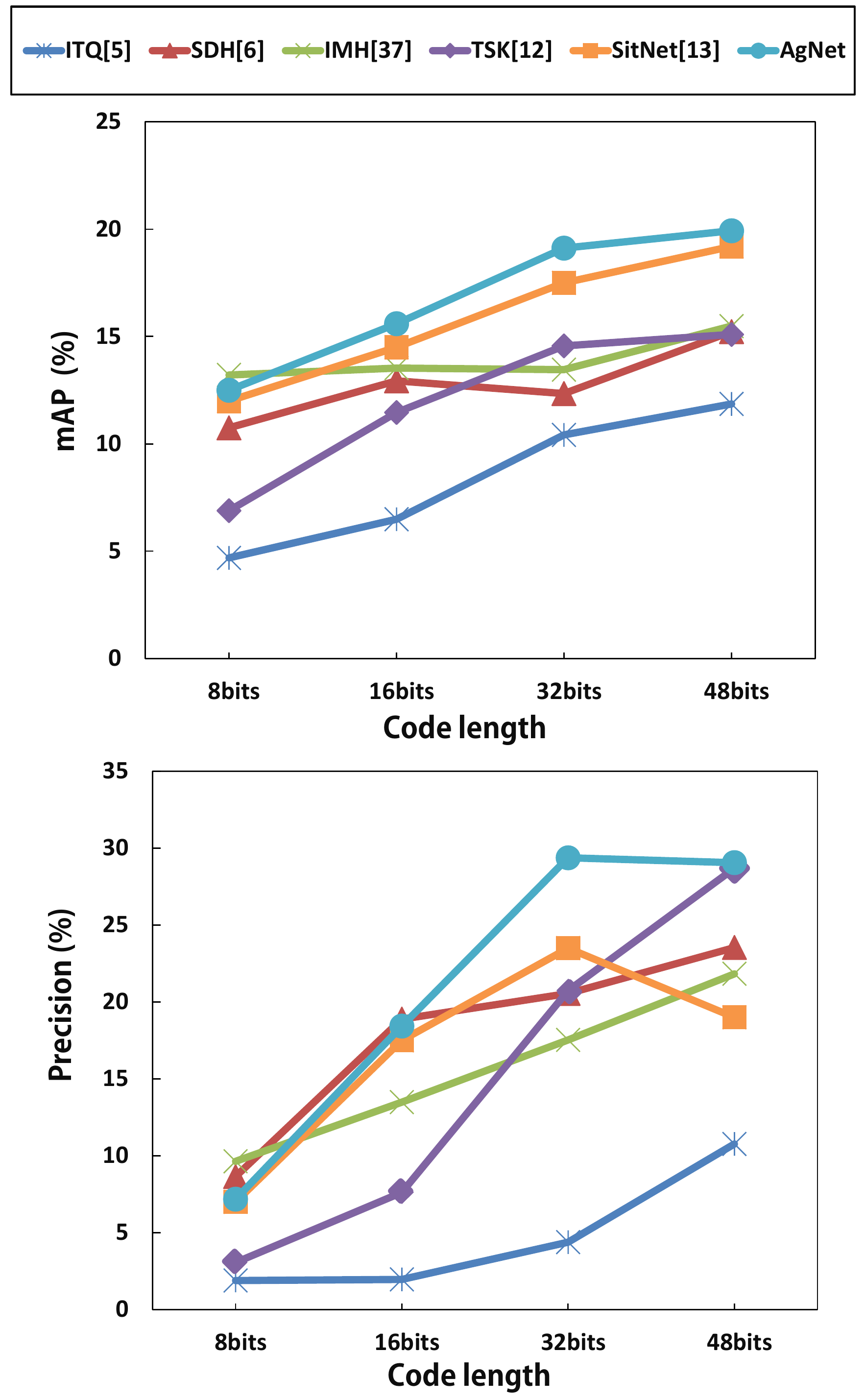}
	\caption{Performances (mAP and Precision) of different methods for single-modal zero-shot hashing task on AwA dataset.}	
\end{figure}
\subsubsection{Experimental results on AwA}All the comparative approaches are implemented by ourselves with the code provided by the authors, except for SitNet. The results of SitNet are directly cited from the original paper \cite{han17ijcai_sitnet}. It should be noted that the split of seen and unseen domain in SitNet has a slight difference with ours. SitNet randomly chooses 10 categories as unseen domain, while we follow the standard split \cite{lampert09cvpr} in this work. We follow this setting to make our work repeatable.
The performances of AgNet and the comparative methods on AwA dataset are reported in Fig. 3. As we can see, the proposed AgNet achieves the best mAP performance in most cases. For example, AgNet gains 19.1\%  in 32 bits, which has an improvement against the second best SitNet by 9.1\% in the same  length. Besides, the mAP performances of AgNet keep improving with the increase of code length, which is similar to the phenomenon in the cross-modal retrieval task. As for Precision, AgNet exceeds all comparative methods in the code length of 32 and 48 bits, and only achieves a slightly inferior performance on 8 and 16 bits. Moreover, there is a slight drop from 32 bits to 48 bits in the precision performance of AgNet, indicating that we need to choose a suitable code length to guarantee the retrieval performance.
\begin{figure}[!h]
	\centering
	
	%	\label{fig:fig3_a}
	\includegraphics[width=0.78\linewidth,height=1.35\linewidth]{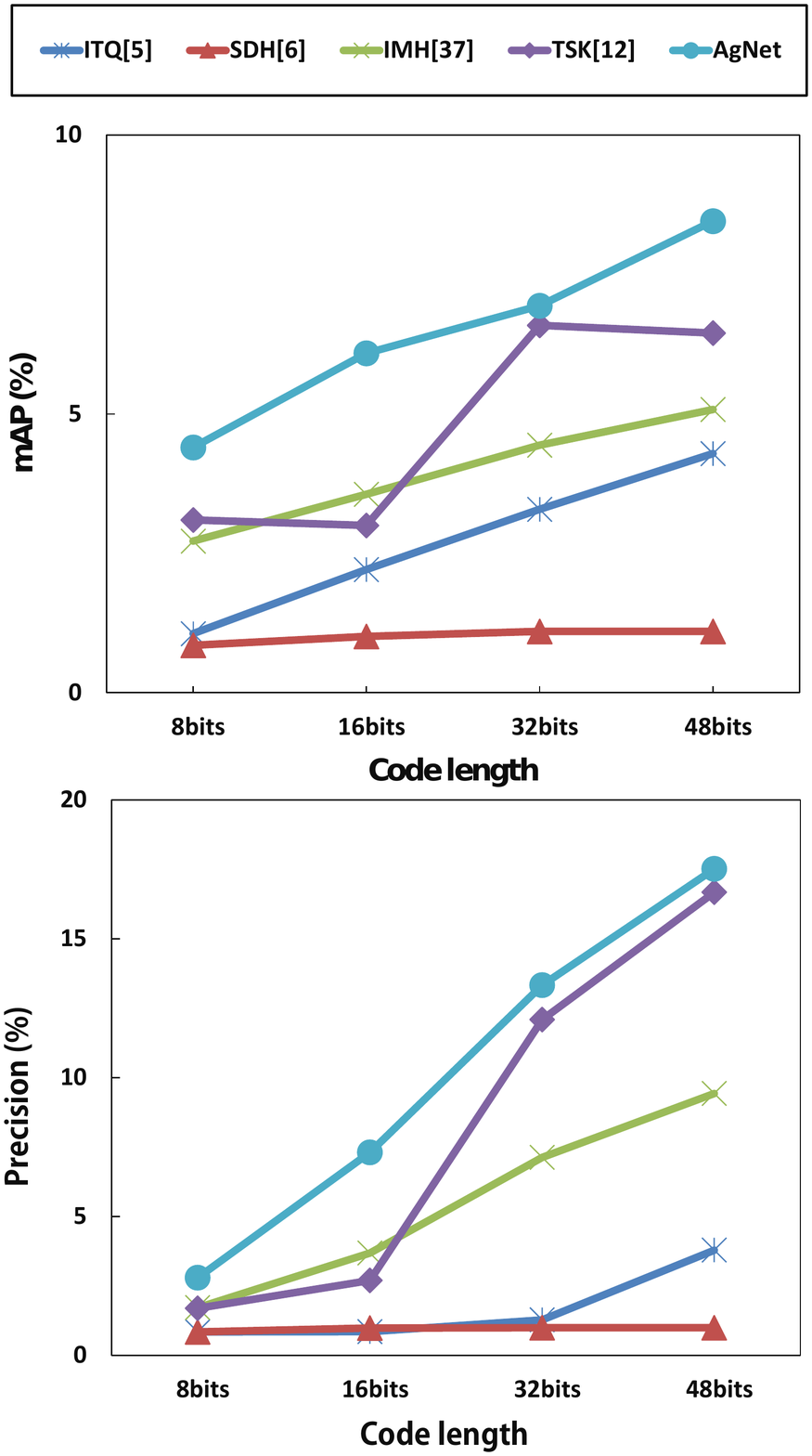}

	\caption{Performances (mAP and Precision) of different methods for single-modal zero-shot hashing task on ImageNet dataset.}	
\end{figure}
\subsubsection{Experimental results on ImageNet}
The comparative experiments are reported in Fig. 4. Note that SitNet \cite{han17ijcai_sitnet} is not selected for comparison in this dataset since its experimental setting is different from ours. It can be observed that AgNet outperforms all comparative methods with significant margins in all code lengths. Besides, the performances of unsupervised methods surpass those of the conventional supervised method, \textit{i.e.}, SDH. Without using the supervision information, unsupervised methods exploit the inherent property of visual representations to generate hash codes and avoid suffering from the misleading of supervision information of seen categories in zero-shot setting. However, by utilizing semantic information as the transferable supervision information, AgNet and TSK mitigate the influence of zero-shot problem and outperform both unsupervised and conventional supervised hashing methods on ImageNet dataset.

\subsection{Effects of Attributes}
As our algorithm is an attribute-based method, the performance of the learned attribute space will affect the discriminative ability of hash codes. In this part, we implement some experiments to analyze the impact of the attribute space, including the scale of attribute space and the attribute prediction accuracy to the final performances.  

To evaluate the influence of attribute space scale, we vary the number of attributes from 10 to 80 with the interval of 10. In consideration of the difference on attributes, we report the average performance of 5 trials for each number by fixing the code length as 64 bits. The curve of CMZSH in terms of mAP is shown in Fig. 5.  

\begin{figure}[!h]
	\centering
	\includegraphics[width=0.65\linewidth, height=0.40\linewidth]{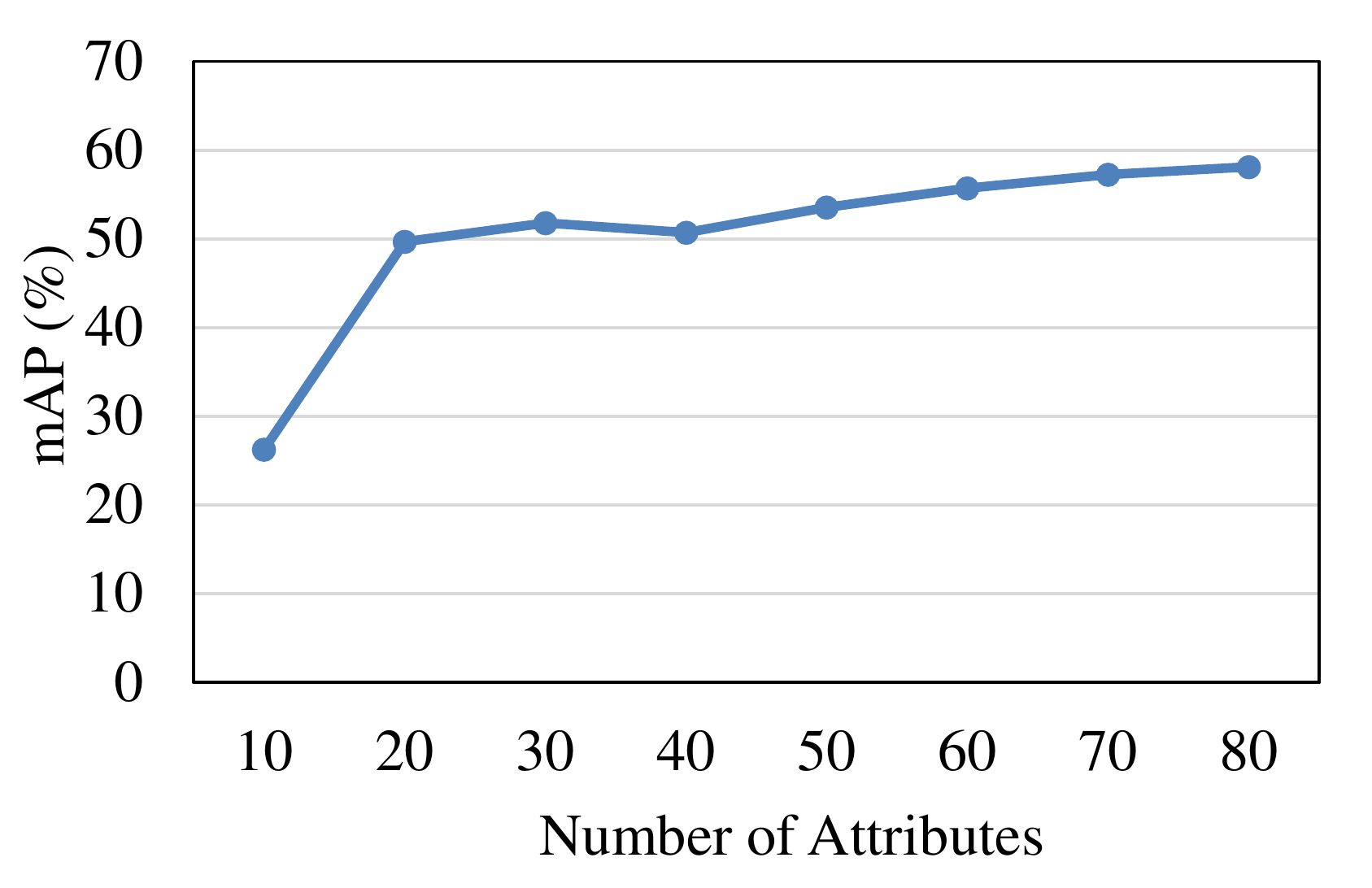}
	\caption{Performances of AgNet with different number of attributes on AwA dataset.}
	
\end{figure}  
It can be observed that the mAP performance increases with the growth of attribute number. Specifically, there is a giant leap when attribute number changes from 10 to 20. It indicates that more attributes are required to guarantee the discriminative ability. Besides, when the amount of attribute is large enough, the increasing scope turns to saturation.

In addition, the attribute prediction accuracy also plays a significant role in the performance of AgNet. The previous experiment in cross-modal task has demonstrated that the performances of AgNet on SUN are inferior to those on AwA. The underlying reason may be that the attribute prediction accuracy on SUN is inferior to those on AwA. Therefore, we analyze the attribute prediction accuracy on both datasets. 
\begin{figure}[!h]
	\centering
	\includegraphics[width=1.0\linewidth]{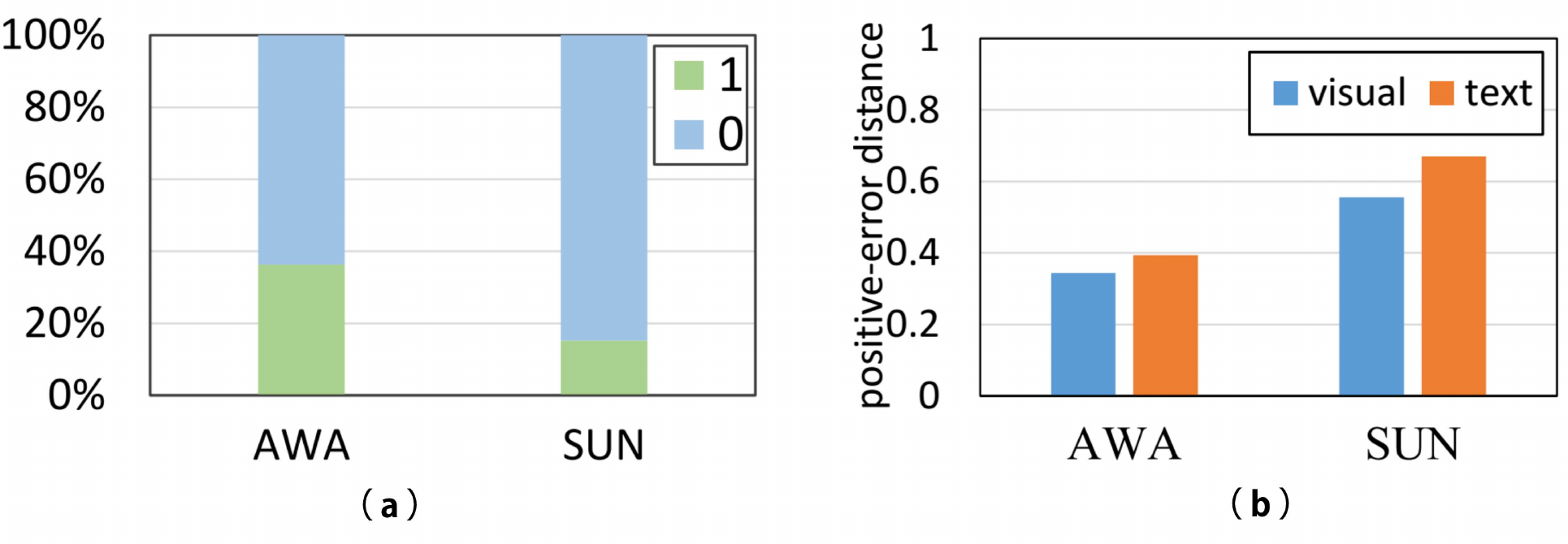}	
	\setlength{\abovecaptionskip}{-0.1cm}
	\caption{(a)The distribution of binary attribute tags on AwA and SUN. (b)The results of positive-error distance on AwA and SUN.}
	
\end{figure}
\begin{figure}[!h]
	\centering
	\includegraphics[width=1.0\linewidth]{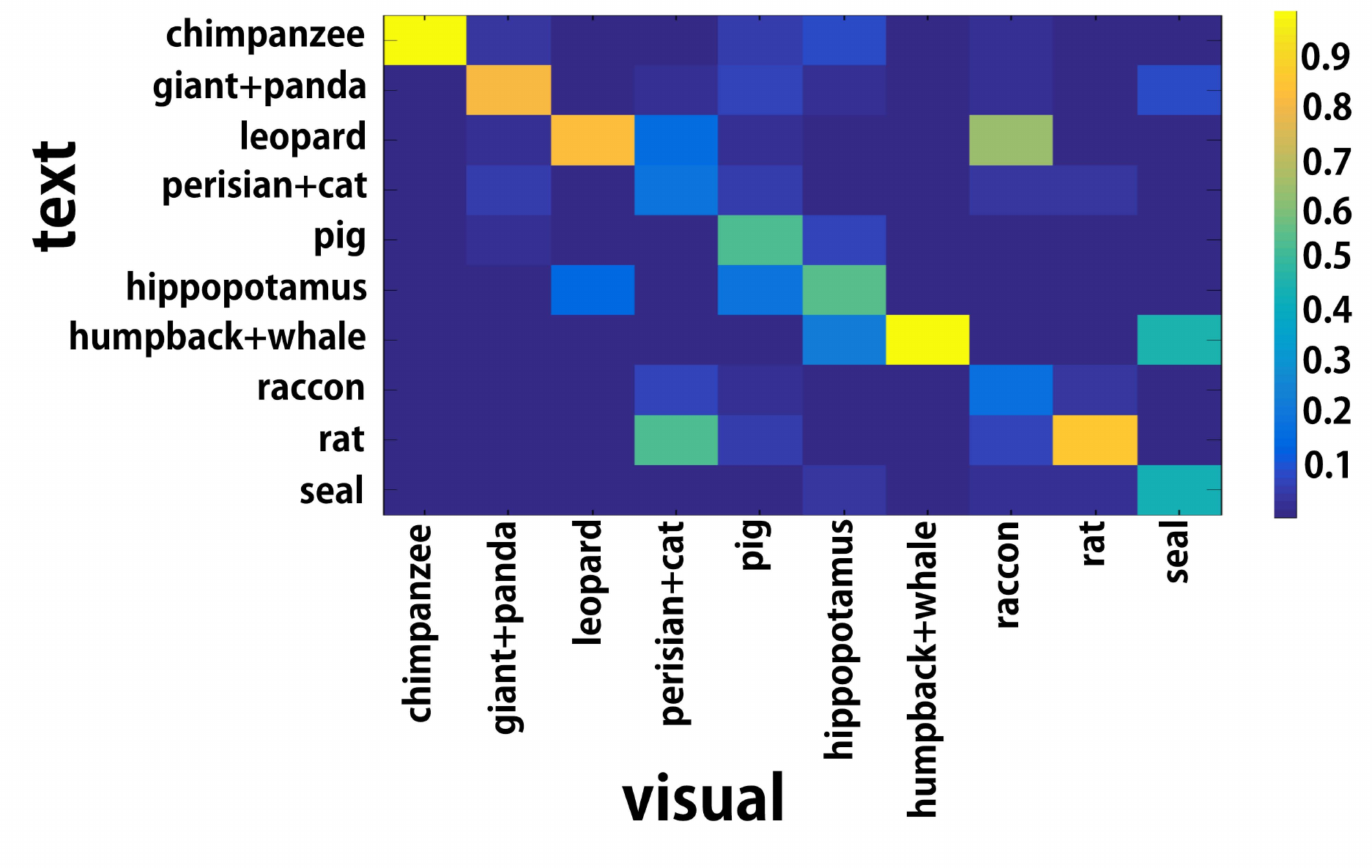}	
	\setlength{\abovecaptionskip}{-0.40cm}
	\setlength{\belowcaptionskip}{-0.51cm}
	\caption{ Confusion matrix of AgNet on AwA, where the columns are the categories that visual hash codes belong to and the rows are the categories of textual hash codes that visual hash codes are close to.}
\end{figure}

According to the distribution of binary attribute tags on AwA and SUN, as is shown in Fig. 6(a), it can be easily noticed that the tags of AwA and SUN are biased to 0.  Therefore, we propose to utilize the positive-error distance  (PED) to evaluate the prediction accuracy, which is defined as:
\begin{equation}
\begin{aligned}
{\cal D} = \frac{{\sum\limits_i^N {\sum\limits_j^d {{\mathbf{A}_{ji}}\left| {{\mathbf{A}_{ji}} - {{\mathbf{\hat A}}_{ji}}} \right|} } }}{{\sum\limits_i^N {\sum\limits_j^d {{\mathbf{A}_{ji}}} } }},
\end{aligned}
\end{equation}
where ${{\mathbf{\hat A}}_{*i}}$ denotes the predicted attribute vector and ${{\mathbf{A}}_{*i}}$ denotes the ground-truth attribute vector, $d$ is the dimensionality of ${{\mathbf{A}}_{*i}}$ and $N$ is the number of instances. Using Eq. (7), the distances between ${{\mathbf{A}}_{*i}}$ and ${{\mathbf{\hat A}}_{*i}}$ are calculated when ${{\bf{A}}_{ji}} = 1$.

The results are reported in Fig. 6(b), which demonstrate that the attribute prediction on AWA is closer to the ground truth than that on SUN. The PED of visual modality and textual modality on SUN are larger than that on AwA in 70.1\% and 61.4\%,  respectively. Thereby, the attribute prediction on AwA is more discriminable than that on SUN, which further interprets the better performance of AgNet on AwA than that on SUN.

\subsection{Visualization}
To further evaluate the performance of AgNet in each category, taking AwA dataset for example, we utilize confusion matrix to visualize the neighbor relationship between textual hash codes and visual hash codes of AgNet. We fix the code length to 64 bits. The result is shown in Fig. 7, where each column denotes the categories that visual instances belong to, and each row is the categories of textual instances that visual instances are close to. It can be observed that most instances are concentrated in the diagonal line, which indicates that visual instances are close to the text instance with the same category in most situations. However, there still exists some confusions in some categories. Take ``seal'' as example, about 40\% of visual instances are close to ``humpback whale''. The main underlying reason is that both categories are marine animal with a lot of similar attributes, which misguides the model to generate the similar hash codes for both categories. This means that the performance of AgNet in similar categories should be further improved in future.
\begin{figure*}[!t]
	\centering
	\includegraphics[width=0.65\linewidth, height=0.25\linewidth]{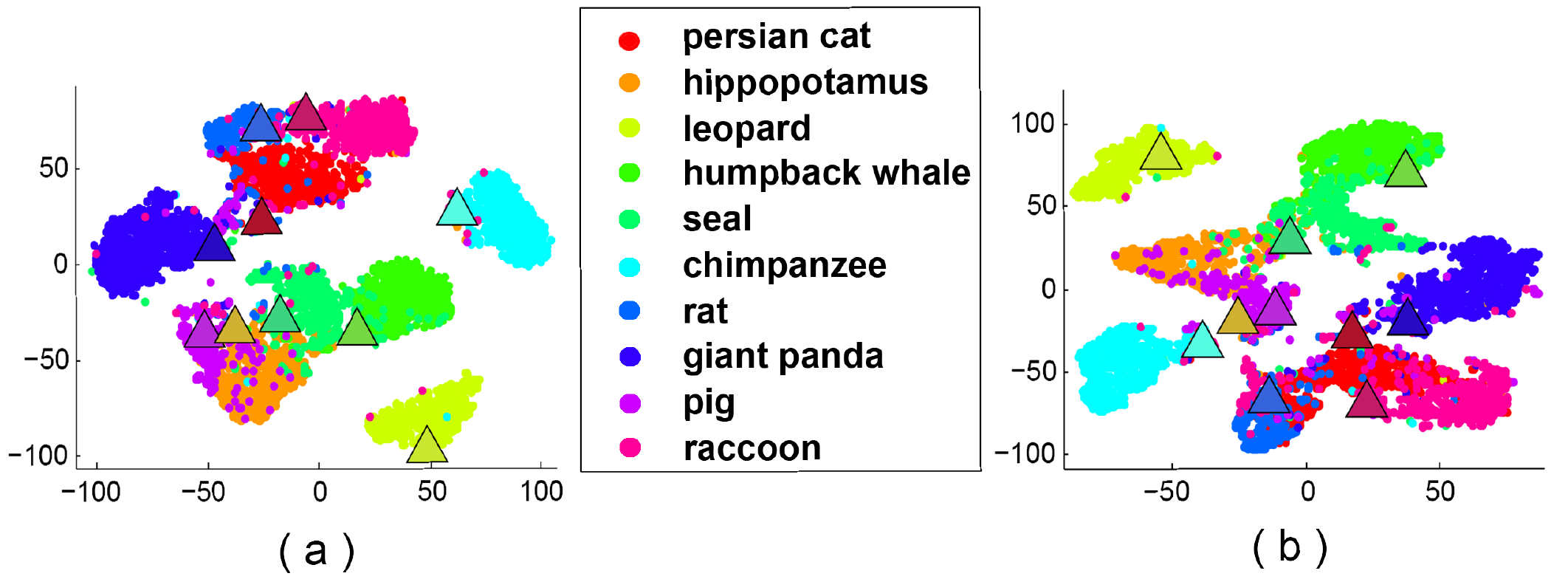}	
	\setlength{\abovecaptionskip}{-0.1cm}
	\caption{ t-SNE visualization of unseen instances on AWA dataset. Points denote visual representations and triangles denote text representations. (a) The visualization of attribute predictions. (b) The visualization of outputs from the last layer in A2H Net.}	
\end{figure*}

In addition, the effective hash codes need to preserve the neighbor relationship of the original features. As for AgNet, we use A2H Net to generate hash codes from both the textual and visual modalities. In this part, we use t-SNE \cite{tsne} to visualize the performance of A2H Net on the unseen domain. Instead of adopting the binary codes that are difficult to generate effective cluster with t-SNE, we utilize attribute predictions and outputs from the last layer in A2H Net as the inputs for t-SNE. As is illustrated in Fig. 8, we can observe that the similarity relationship in attribute space has been well preserved in the hash space, which indicates the effectiveness of A2H Net.
\section{Conclusion}
In this paper, we have proposed a deep hashing neural network to address the cross-modal zero-shot retrieval problem. It aligns different modal data into a more high-level semantic space, \textit{i.e.}, attribute space. Besides, category similarity is utilized to construct the relationships between different modalities while attribute similarity is introduced to regularize the distance of similar categories in single modality. Experimental results on both cross-modal and single-modal retrieval tasks have demonstrated the superiority of the proposed approach.

In the future, as the acquisition of attribute annotation requires prior knowledge, we shall exploit other semantic data to formulate common space, \textit{e.g.}, click-through data.
%\appendices 

\ifCLASSOPTIONcaptionsoff
  \newpage
\fi


%\end{IEEEbiography}


%\newpage



%\vfill


\end{document}